\documentclass[runningheads]{llncs}
\usepackage{wrapfig}
\usepackage[utf8]{inputenc} 
\usepackage[T1]{fontenc}    
\usepackage{hyperref}       
\usepackage{url}            
\usepackage{booktabs}       
\usepackage{amsfonts}       
\usepackage{nicefrac}       
\usepackage{microtype}      
\usepackage{xcolor}         
\usepackage{amssymb}
\usepackage{enumitem}
\usepackage[pdftex]{graphicx}
\usepackage{comment}
\usepackage{multirow}   
\usepackage{colortbl}   
\usepackage{booktabs}   
\usepackage{makecell}
\usepackage{tcolorbox}
\usepackage[normalem]{ulem}
\usepackage{CJKutf8}
\usepackage{fontawesome}
\usepackage{graphicx}
\usepackage{multirow}

\begin{document}
%
\title{Are MLMs Trapped in the Visual Room?}
\author{
Yazhou Zhang\inst{2,3}, Chunwang Zou\inst{1} \and Qimeng Liu\inst{1} \and Lu Rong\inst{2} \and Ben Yao\inst{3} \and Zheng Lian\inst{4,*} \and Qiuchi Li\inst{5,*} \and Peng Zhang \inst{2} 
\and Jing Qin\inst{3}}
\authorrunning{Zhang et al.}
%
\institute{Zhengzhou University of Light Industry \and
Tianjin University \and The Hong Kong Polytechnic University \and Institute of Automation, Chinese Academy of Sciences \and Beijing Institute of Technology\\
\email{Correspondence: yzhou\_zhang@tju.edu.cn, lianzheng2016@ia.ac.cn, liqiuchi2015@gmail.com}}
%
\maketitle              
\begin{abstract}
Can multi-modal large models (MLMs) that can ``see'' an image be said to ``understand'' it? Drawing inspiration from Searle's Chinese Room, we propose the \textbf{Visual Room} argument: a system may process and describe every detail of visual inputs by following algorithmic rules, without genuinely comprehending the underlying intention. This dilemma challenges the prevailing assumption that perceptual mastery implies genuine understanding.
In implementation, we introduce a two-tier evaluation framework spanning perception and cognition.
The perception component evaluates whether MLMs can accurately capture the surface-level details of visual contents, where the cognitive component examines their ability to infer sarcasm polarity. 
To support this framework, We further introduce a high-quality multi-modal sarcasm dataset comprising both 924 static images and 100 dynamic videos. All sarcasm labels are annotated by the original authors and verified by independent reviewers to ensure clarity and consistency. 
We evaluate eight state-of-the-art (SoTA) MLMs. Our results highlight three key findings: (1) MLMs demonstrate high accuracy in visual perception; (2) even with correct perception, MLMs exhibit an average error rate of ~17.1\% in sarcasm understanding, revealing a significant gap between seeing and understanding; (3) this gap stems from weaknesses in context integration, emotional reasoning, and pragmatic inference.
This work provides empirical grounding for the proposed Visual Room argument and offers a new evaluation paradigm for MLMs.
\end{abstract}
\section{Introduction}
Large language models (LLMs) have swept across natural language processing (NLP). Due to the multi-modal nature of human cognition, this momentum has naturally extended into the multi-modal domain. Through continued pretraining and instruction tuning on large-scale image-text pair datasets, a wave of MLMs such as  GPT-4V~\cite{wu2023can}, LLaVA 1.5~\cite{liu2024improved}, Qwen 3-VL~\cite{bai2025qwen2}, etc., has emerged. Such MLMs have demonstrated impressive performance on visual perception tasks, e.g., image classification, object detection and segmentation, proving their strong perceptual capabilities. 

However, $\mathrm{seeing}  \ne \mathrm{understanding} $. As John Searle's Chinese Room thought experiment reveals that symbol manipulation does not equate to semantic understanding~\cite{searle2009chinese}, we further propose the \textbf{Visual Room} argument (VRA):
\begin{figure}[t]
  \centering
  \includegraphics[width=3.5in]{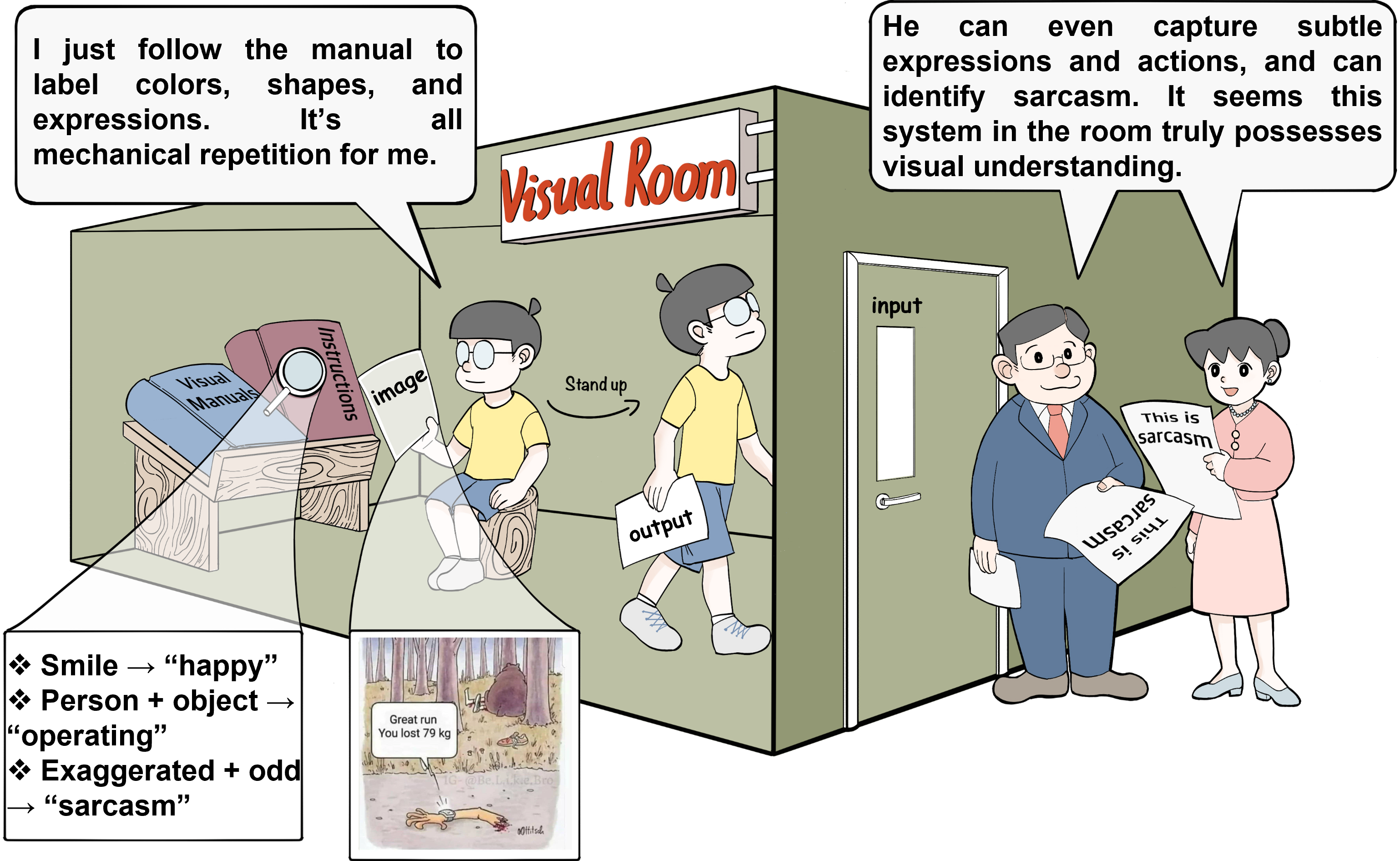}
  \caption{The proposed Visual Room argument.}
  \label{fig:vra}
\end{figure}

\begin{itshape}
Imagine a scenario where a visual symbol operator with no understanding of visual semantics is locked inside a sealed room with only a small opening. The room is filled with manuals containing rules for describing visual features such as shapes, colors, objects, and facial expressions, as well as an instruction book (e.g., a MLM) that details how to process these features. Whenever an image is passed into the room through the opening (input), the operator inside mechanically follows the instructions: (1) meticulously records all visual elements in the image, such as facial expressions, objects, and scene composition (scene recognition); (2) combines these symbolized visual details into a descriptive text, or makes a judgment: ``this is sarcasm'' or ``this is not sarcasm'' (output).

To an outside observer, this Visual Room seems capable of recognizing and describing complex images, and even correctly judging non-literal intentions such as sarcasm or metaphor. In reality, however, the operator inside has no genuine understanding of the image's true intent, and just simply manipulating symbols according to a set of rules.
\end{itshape}

The perception–cognition gap derived from this VRA (see Fig.~\ref{fig:vra}) is central to judging whether current MLMs are truly progressing toward human-like general intelligence.
Multi-modal sarcasm understanding, as a subtle and non-literal form of intentional expression, offers a sharp lens through which to examine this gap. Correctly interpreting sarcasm requires not only precise perception of visual content, but also higher-order cognitive mechanisms, such as emotion recognition, contextual reasoning, etc. 
Hence, sarcasm understanding may serve as a ``stress test'' for evaluating the perception–cognition gap.

To operationalize and quantify the proposed VRA, we propose a universal ``perception-cognition'' two-tier evaluation framework. On the perception level, we provide human-authored scene descriptions to test whether the MLMs can accurately ground visual semantics. On the cognitive level, we evaluate the MLMs' ability to recognize and interpret sarcastic polarity. 

In view that existing sarcasm datasets (e.g., CMMA~\cite{zhang2023cmma}, MUStARD~\cite{castro2019towards}) only annotate sarcasm labels and lack fine-grained scene descriptions, sarcasm detection models suffer from a disconnect between perception and cognition. To address this issue, we further construct a high-quality multi-modal sarcasm dataset comprising both 924 static images and 100 dynamic videos. Unlike prior work, all sarcasm labels are annotated by the content authors themselves and verified by multiple independent reviewers to ensure accuracy and interpretability. Additionally, scene descriptions are provided for each sample to make the MLMs' perceptual and cognitive performance explicitly measurable.

Grounded in this framework, we evaluate six state-of-the-art (SoTA) MLMs (i.e., Claude 3.7 Sonnet, GPT-4V, Gemini 2.0 Flash, DeepSeek VL, Qwen VL-Plus, and GLM 4V-Plus) through four prompting approaches (including ours). Our results highlight three key findings: (1) MLMs demonstrate high accuracy in visual perception, particularly for static images; (2) even with correct perception, MLMs exhibit an average error rate of ~17.1\% in sarcasm understanding, revealing a significant gap between seeing and understanding; (3) this gap stems from weaknesses in context integration, emotional reasoning, and pragmatic inference. Our main contributions are written as:
\begin{itemize}
    \item We propose the Visual Room argument, offering a new theoretical lens to formalize the perception–cognition gap in MLMs.
   \item   We design a universal two-tier evaluation framework to disentangle visual perception from cognitive understanding.
    \item We introduce a high-quality multi-modal sarcasm dataset with author-annotated and reviewer-verified labels, offering a solid testing ground for assessing the cognitive capabilities of MLMs. 
\end{itemize}

\section{Related Work}
\subsection{Multi-Modal Large Models}
MLMs have emerged as a convergence of LLMs and VLMs, aiming to unify perception and reasoning across modalities. Early works such as CLIP~\cite{radford2021learning} leveraged large-scale image–text pairs and contrastive learning to establish foundational visual-textual alignment. Subsequent models, including Flamingo~\cite{alayrac2022flamingo} and BLIP-2~\cite{li2023blip}, introduced learnable vision-language adapters, enabling effective multi-modal fusion. 
More recently, LLM-based multi-modal models, e.g., GPT-4V, LLaVA~\cite{liu2024improved}, and MiniGPT-4~\cite{zhu2023minigpt}, have advanced this line of work by incorporating carefully designed modality encoders and adapters, and have been trained on massive image-text paired corpora to learn world knowledge. 
These models have achieved strong results on image captioning, visual question answeringm and so on, highlighting their impressive perceptual capabilities. 

However, current MLMs predominantly focus on perception-level tasks, i.e., whether the model can ``see'' and describe visual content. Whether MLMs can truly understand implied meaning remains underexplored.

\subsection{Multi-Modal Sarcasm Understanding}
Prior studies on the cognitive abilities of MLMs typically followed two directions: commonsense reasoning (e.g., Miko~\cite{lu2024miko}) and emotion recognition (e.g., Emotion-LLaMA~\cite{cheng2024emotion}, DialogueLLM~\cite{zhang2023dialoguellm}). They primarily targeted explicit semantic understanding while overlooking the complexity of non-literal meanings.
In contrast, sarcasm understanding represents a prototypical higher-order cognitive task that demands models to move beyond literal interpretation toward pragmatic inference. A number of multi-modal benchmarks have been proposed, such as CMMA~\cite{zhang2023cmma}, MUStARD~\cite{castro2019towards}, and MMSD~\cite{qin2023mmsd2}. Such benchmarks are typically constructed by collecting image/video–text pairs from social media platforms or scripted sitcoms, followed by human annotation of sarcasm polarity. However, they are limited by their binary sarcasm annotations and dependence on third-party annotators, whose inconsistent understanding of context and author intent often leads to unreliable and noisy labels. In addition, they lack rich scene descriptions, which prevents subsequent sarcasm recognition approaches, e.g., MuMu~\cite{wang2024cross}, etc., from disentangling perceptual representation from cognitive inference. As a result, nearly all SoTA models treat sarcasm understanding as a binary classification task, without investigating whether failures stem from deficits in low-level perception (e.g., missing visual cues) or higher-order cognition (e.g., pragmatic inference).

The above-mentioned limitations highlight a critical gap in evaluating the gap between perception and cognition. There is an urgent need for a sarcasm understanding framework that is both semantically rich and diagnostically precise. This work represents a concrete step toward that goal.

\section{Methodology}
\begin{figure}[t]
  \centering
  \includegraphics[width=4.8in]{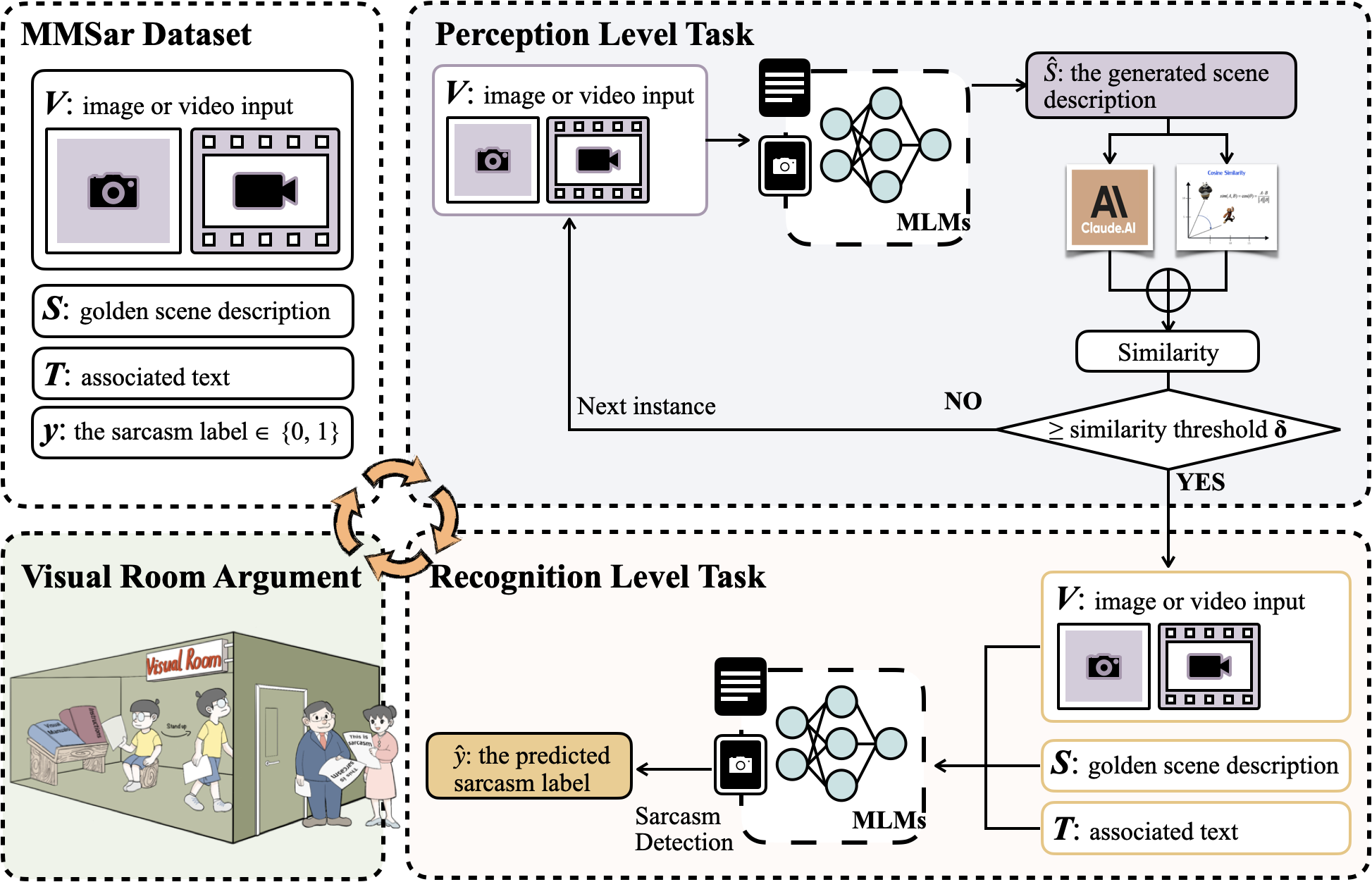}
  \caption{The proposed perception-cognition framework.}
  \label{fig:framework}
\end{figure}

\subsection{Visual Room Argument}
The Visual Room thought experiment is inspired by John Searle's Chinese Room argument, which contrasts external appearances of language understanding with the internal reality of mere symbol manipulation. In contrast, the Visual Room extends this dilemma from the domain of language to the multi-modal realm, focusing on whether MLMs can truly bridge the gap from mechanical perception to genuine understanding, especially in high-level cognitive tasks such as pragmatic reasoning, sarcasm understanding.

The essence of the Visual Room lies in revealing that, even if a system can flawlessly decompose and describe every visual detail in an image (such as objects, facial expressions, and scene composition) according to prescribed rules, its ``understanding'' may remain confined to mechanical symbol processing, never reaching the underlying context or intent. This experiment echoes the current challenge faced by MLMs: excelling in perception, yet struggling with higher-order cognitive tasks. By concretizing this thought experiment, we aim to delineate the boundary between perception and cognition in model capabilities.

Unlike purely thought experiments, the Visual Room is intended as a transferable and operational theoretical paradigm: as long as we design appropriate layered tasks and provide systematic diagnostics, this logic can be applied and validated across a wide range of multi-modal challenges. Even with other visual tasks (such as metaphor recognition), the Visual Room remains robust wherever symbol manipulation is decoupled from intent understanding.


\subsection{The Perception–Cognition Framework}
Building on the Visual Room argument, We propose a two-tier perception–cognition evaluation framework to explicitly disentangle perception from cognition, as shown in Fig.~\ref{fig:framework}. 

Support that each instance is represented as a quadruple: $x = \{V, T, S, y\}$, where $V$ is the image or video input, $T$ is the associated text, $S$ is the golden scene description for $V$, and $y \in \{0,1\}$ is the sarcasm label.
The overall objective is written as:
\begin{equation}
P(y, S | V, T) = P(y | S, V, T) \cdot P(S | V)
\end{equation}
where
$P(S|V)$ corresponds to the \textbf{perception task}. In contrast, $P(y | S, V, T)$ corresponds to the \textbf{cognition task}.

\textbf{Perception level.} The perception task aims to estimate $P(S | V)$. In practice, the model generates a visual description $\hat{S}$, which is then compared against a human-authored reference description $S$. We define the perception correctness as:
\begin{equation}
\mathrm{Acc} _{P}=\mathbb{I}\left [ sim\left ( \hat{S},S  \right ) \ge \delta  \right ]  
\end{equation}

To ensure evaluation robustness, we employ a hybrid approach combining cosine similarity and third-party LLM (namely Claude 3.7) judgment:
\begin{equation}
sim(\hat{S}, S) = \frac{1}{2}  \cdot sim_{cos}(\hat{S}, S) + \frac{1}{2}  \cdot sim_{llm}(\hat{S}, S)
\end{equation}


We set the similarity threshold $\delta=0.8$ for image samples and $\delta=0.65$ for video samples, based on the results of our pilot study. A sample is considered perceptually grounded if $\mathrm{Acc} _{P}=1$.

\textbf{Cognition level.} On perceptually grounded samples, we further assess the model's ability to predict sarcastic polarity, i.e., estimating $P(y | S,V, T)$. The cognition-level accuracy is defined as:
\begin{equation}
\mathrm{Acc} _C = \mathbb{P}(\hat{y} = y | \mathrm{Acc} _{P}=1)
\end{equation}
which measures the model's capacity for sarcasm inference under correct perceptual grounding. 

Hence, to quantify the sarcasm understanding failure even when perception succeeds, we further define the \textit{perception–cognition gap}:
\begin{equation}
\mathrm{Gap}_{P, C} = \mathbb{P}(\hat{y} \neq y |\mathrm{Acc} _{P}=1)
\end{equation}

By disentangling errors into perception and cognition layers, our framework moves beyond surface-level accuracy and provides interpretable insight into \textit{why} a model succeeds or fails, and \textit{whether} it misperceives visual input or misinterprets pragmatic intent. 

\subsection{Multi-Modal Sarcasm Dataset Construction}
Existing multi-modal sarcasm datasets (e.g., CMMA~\cite{zhang2023cmma}, MUStARD~\cite{castro2019towards}, MMSD~\cite{qin2023mmsd2}) suffer from two major limitations. (1) they typically annotate only sarcasm polarity, lacking fine-grained semantic descriptions of the scene, which disconnects the evaluation of perception and cognition. (2) they rely heavily on third-party crowdsourced annotations, whose quality varies with annotators' expertise, interpretive accuracy, and alignment with the author's original intent. This introduces substantial noise and undermines label reliability. Hence, to support the proposed framework, we construct a high-quality multi-modal sarcasm dataset (MMSar), with a primary focus on the reliability of sarcasm annotations. 

\textbf{Data acquisition.}
We select Reddit as our primary source, due to its structured, topic-centric architecture. Unlike platforms such as Facebook, Instagram or Weibo, Reddit organizes user interactions into thousands of topical subreddits, each of which governed by its own topic. This structure allows for precise targeting of sarcasm-rich content within clearly defined discourse contexts.

To this end, we extract posts from three dedicated sarcasm-related subreddits, i.e., \textit{r/sarcasm}, \textit{r/irony}, and \textit{r/satire}, where sarcastic expression is not only prevalent but thematically central. To ensure interpretability and reliability, we define three selection criteria: (1) each post must include an explicit author-provided tag such as ``sarcasm'' or ``irony'', ensuring that the sarcastic intent is self-annotated rather than externally inferred; (2) each post must receive a minimum of three comments or five upvotes, serving as weak supervision that the sarcastic meaning has been recognized and socially affirmed; (3) posts must contain both visual contents (image or video) and accompanying texts.

To construct a balanced dataset, we adopt the same sampling strategy for non-sarcastic posts, sourcing from thematically distinct subreddits such as `` r/love'',`` r/cute'', ``r/happy'', etc. These posts are also labeled by the original authors, ensuring that both classes exhibit comparable structure and community coherence. As a result, we collect an initial pool of 1,057 candidate multi-modal samples.

\textbf{Quality control.} We enforce strict quality control through a multi-stage screening process. An expert reviews all candidate samples and eliminates any with ambiguous sarcasm intent. We further remove samples with low-resolution or unclear images, and any instances with vague rhetorical targets.

\textbf{Fine-grained scene descriptions.} To evaluate the model's visual perception capability, we provide a high-quality human-written scene description for each sample. These descriptions aim to faithfully capture the visual content of the image or video. Specifically, we recruit three annotators (one master student and two undergraduates) to complete the annotation. For each sample, two annotators independently write a scene description without access to the paired text, and the third annotator perform review and refinement. To standardize the annotation process, we design an annotation guideline: (1) cover all salient visual elements, e.g., people, objects, facial expressions, actions, and background context; (2) describe the spatial layout and interactions among elements; (3) avoid any subjective or emotional interpretation; (4) use third-person, objective language; and (5) keep the description between 50–150 words to ensure completeness without redundancy.

Prior to annotation, all annotators received training based on the annotation guidelines, including definitions of visual elements, common pitfalls, and examples of both acceptable and unacceptable descriptions. We also addressed follow-up questions raised by the annotators regarding the instructions. To ensure consistency, they were first asked to annotate 10 pilot samples, with a target inter-annotator agreement of 90\%.  We calculate the Cohen's kappa score, $\kappa =0.83$, which means the two annotators have reached high agreement.
Note that sarcasm labeling requires subjective interpretation of the author’s intent, while scene descriptions are objective records of visible content. With clear guidelines and sufficient training, scene annotation can be performed reliably by human annotators.

Finally, the dataset comprises 1024 high-quality multi-modal samples, including 924 image–text pairs and 100 video–text pairs, with 521 sarcastic and 503 non-sarcastic instances, as shown in Table~\ref{tab:tab1}.
\begin{table}[t]
\centering
\footnotesize     
\caption{Statistics of the MMSar dataset.}
\label{tab:tab1}
\begin{tabular}{c|l|c}
\toprule
\textbf{~~Modality~~}                    & \multicolumn{1}{l|}{\textbf{Item~~}}           & \textbf{~~Num~~}   \\ \hline
\multirow{4}{*}{Images} & \#Sarcastic                        & 471   \\ 
                        & \#Non-Sarcastic                    & 453   \\ 
                        & \#Avgerage text length             & 10.6 \\  
                        & \#Average scene description length & 95.6 \\ \hline
\multirow{5}{*}{Videos} & \#Sarcastic                        & 50    \\
                        & \#Non-Sarcastic                    & 50    \\ 
                        & \#Avgerage text length             & 10.4     \\ 
                        & \#Average scene description length & 129.7     \\ 
                        & \#Average duration of per video    & 43s    \\ 
                        \toprule
\end{tabular}
\end{table}

\begin{figure}[t]
  \centering
  \includegraphics[width=4.9in]{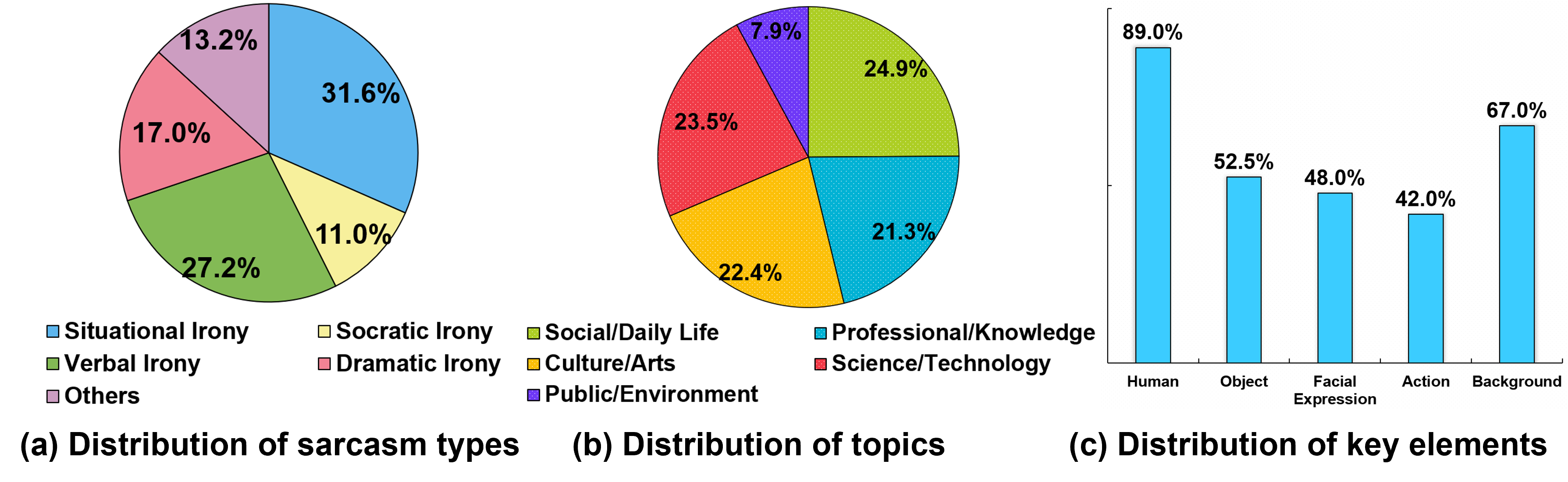}
  \caption{The distribution of sarcasm types, topics, and key visual elements.}
  \label{fig:datasetanalysis}
\end{figure}
\subsection{Dataset Analysis} 
\textbf{Distribution of sarcasm types.} We conduct a statistical analysis of sarcasm types, as shown in Fig.~\ref{fig:datasetanalysis} (a). The results show that situational irony accounts for the largest proportion at 31.6\%, followed by verbal irony at 27.2\%, dramatic irony at 17.0\%, Socratic irony at 11.0\%, and other types at 13.2\%. This distribution demonstrates that the dataset not only covers mainstream forms of sarcasm but also includes a diverse range of pragmatic scenarios, providing a solid foundation for evaluating the multidimensional cognitive abilities of MLMs.

\textbf{Topic analysis.} We categorize all samples into five topic groups in Fig.~\ref{fig:datasetanalysis} (b). The largest category is Social/Daily Life (24.9\%) and includes topics such as family, health, psychology, and travel, demonstrating that sarcasm frequently arises from everyday experiences and personal interactions. The next three categories, i.e., Science/Technology (23.5\%), Culture/Arts (22.4\%), and Professional/Knowledge topics (21.3\%), are of comparable size, together reflecting the dataset's strong representation of scientific discourse, cultural expression, and specialized knowledge domains. This balance ensures that models evaluated on this dataset can generalize across a wide range of real-world scenarios. Finally, Public/Environment account for 7.9\%, capturing societal, environmental, and sports-related sarcasm. Overall, this distribution underscores the dataset's diversity and comprehensive topical coverage, making it well-suited for sarcasm understanding evaluation.

\textbf{Analysis of scene descriptions.} We systematically analyze the coverage of key visual elements within the scene descriptions in Fig.~\ref{fig:datasetanalysis} (c).  The results show that 89.0\% of descriptions mention humans, making them the most prominent element in the dataset. Objects are referenced in 52.5\% of cases, followed by background details at 67.0\%, reflecting an emphasis on both central figures and contextual information. Additionally, 48.0\% of descriptions include explicit references to facial expressions, while 42.0\% capture specific actions. This distribution demonstrates that the scene descriptions offer rich and multi-dimensional visual information, providing a robust foundation for evaluating models' visual perception capabilities.

\section{Experiments}
Our experiments serve as the empirical cornerstone to validate the Visual Room argument, operationalize the perception–cognition evaluation framework, and demonstrate the broader applicability of our dataset.

\subsection{Experiment Setups}\label{sec:setups}

\textbf{Baselines.}  
We conduct evaluation experiments on MMSar over 8 SoTA MLMs via four prompting methods. They are: (1) \textbf{\underline{Claude 3.7 Sonnet}}\footnote{https://claude.ai/}, (2) \textbf{\underline{GPT-4V}}~\cite{hurst2024gpt}, (3) \textbf{\underline{Gemini-2.0-Flash}}\footnote{https://gemini.google.com/app}, (4) \textbf{\underline{DeepSeek-VL}}~\cite{lu2024deepseek}, (5) \textbf{\underline{Qwen-VL-Plus}}~\cite{wang2024qwen2}, (6) \textbf{\underline{GLM-4V-Plus}}~\cite{glm2024chatglm}, (7) \textbf{\underline{LongVU}}~\cite{shen2024longvu} and (8) \textbf{\underline{Chat-UniVi}}~\cite{jin2024chat}.

\textbf{Prompting methods.} They are: (1) zero-shot Input/Output (IO) Prompting, (2) multi-modal chain of thought (MCoT)~\cite{zhang2024multimodalchainofthoughtreasoninglanguage}, and (3) generated knowledge (GK) prompting~\cite{liu2021generated}, and (4) our proposed scene-augmented IO (SA-IO) prompting, which incorporates explicit scene descriptions into the IO prompting for enhanced perception–cognition alignment. They are used to elicit different reasoning patterns from the MLMs.

\begin{table}[t]
\centering
\footnotesize     
\caption{Comparison of 8 MLMs on MMSar. \textbf{Bold} indicates the best and \underline{underline} the second.}
\label{tab:baseline}
\begin{tabular}{llcccccccc}
\toprule
\multirow{2}{*}{\textbf{Modality}} & \multirow{2}{*}{\textbf{Model}} & \multicolumn{2}{c}{\textbf{IO}} & \multicolumn{2}{c}{\textbf{MCoT}} & \multicolumn{2}{c}{\textbf{GK}} & \multicolumn{2}{c}{\textbf{SA-IO}} \\ \cline{3-10}
                             &  & ~Acc.~            & ~F1~            &~Acc.~   & ~F1~                        & ~Acc.~                    & ~F1~                     & ~Acc.~                & ~F1~              \\ \toprule
\multirow{6}{*}{\textbf{Image}} & \textbf{GPT-4V }                       & \textbf{85.6}                 & \textbf{86.7 }              & 82.0       & 84.2                           & 76.3                         & 80.1                       & 81.0                    & 82.8                 \\
                              
& \textbf{Claude-3.7-Sonnet }            & \underline{85.4}                 & \underline{85.9}               & 78.1       & 78.7                           & 77.8                         & 77.0                       & 85.0                    & 85.5                 \\
&\textbf{Qwen-VL-Plus}                  & 81.3                 & 80.5               & 75.8       & 73.6                           & 56.1                         & 65.0                       & 83.0                    & 82.4                 \\
& \textbf{Gemini-2.0-Flash}          & 82.0                 & 82.6               & 77.5       & 76.5                           & 76.8                         & 73.0                       & 77.8                    & 79.6                 \\

& \textbf{DeepSeek-VL}                   & 65.2                 & 51.9               & 62.9       & 48.7   & 58.8                         & 55.3                       & 76.2                    & 72.4                 \\
& \textbf{GLM-4V-Plus}                   & 79.5                 & 82.1               & 71.1       & 74.5                           & 74.1                         & 74.1                       & 80.0                    & 82.4                 \\ \toprule
\multirow{2}{*}{\textbf{Video}} & \textbf{LongVU} & 45.8 & 10.3 & \textbf{69.1} & \textbf{67.5} & 64.9 & 49.2 & 40.6 & 3.4 \\
& \textbf{Chat-UniVi} & 52.6 & 60.0 & 55.1 & \underline{67.2} & 58.5 & \underline{65.6} & 47.5 & 56.7 \\
\toprule
\end{tabular}
\end{table}

\subsection{End-to-End (Single-Stage) Evaluation Results} \label{surface Results}
We begin by presenting the results of the standard end-to-end (single-stage) evaluation, where MLMs are required to directly classify each instance as sarcastic or non-sarcastic without explicit separation of perception and cognition.  We report both \textbf{Accuracy} and \textbf{F1} results in Table~\ref{tab:baseline}. The results show that GPT-4V achieves the best performance under the zero-shot IO setting (86.7 F1), with Claude-3.7-Sonnet as a close second (85.9 F1). Qwen-VL-Plus and Gemini-2.0-Flash perform moderately well, while DeepSeek-VL shows the lowest baseline results.

The MCoT and GK prompting methods do not consistently improve sarcasm understanding performance. For example, GPT-4V's F1 decreases by 4.2\% from IO to MCoT, and by 8.5\% for Qwen-VL-Plus (from 80.5 to 73.6). The reason is that sarcasm understanding is often considered a holistic and non-rational cognitive process that does not conform to step-by-step logical reasoning. Complex reasoning prompts may introduce noise or cognitive overload.

When we augment the input with explicit scene descriptions (namely SA-IO), the results are mixed. DeepSeek-VL benefits most, with its accuracy increasing by 39.5\% (from 51.9 to 72.4), indicating that additional perceptual cues help weaker MLMs. GLM-4V-Plus and Claude 3.7-Sonnet show no significant change in performance after incorporating scene descriptions.
In contrast, GPT-4V's F1 drops by 4.5\%, and Gemini-2.0-Flash drops by 4.1\%. This shows that additional scene descriptions may introduce redundant or conflicting information into these stronger MLMs, leading to confusion rather than improved understanding. For video-based tasks, both LongVU and Chat-UniVi achieve notably lower performance than image-based models, indicating that sarcasm understanding in dynamic visual contexts remains even more challenging.

\uline{These findings empirically support our Visual Room argument: mastery of surface-level perception, even when models are given the ground-truth scene descriptions, does not necessarily translate to better cognitive comprehension.} In fact, supplementing MLMs with explicit perceptual information may expose their inability to integrate perception and cognition. This persistent gap highlights the fundamental limitations of current MLMs in bridging the perception–comprehension divide.

\begin{table}[t]
\centering
\footnotesize     
\caption{The \textbf{perception} capability evaluation results of 8 MLMs.}
\label{tab:perception}
\begin{tabular}{llcccc}
\toprule
\multirow{2}{*}{\textbf{Modality}} & \multirow{2}{*}{\textbf{Model}} & \multicolumn{2}{c}{\textbf{IO}} & \multicolumn{2}{c}{\textbf{MCOT}} \\
\cmidrule(lr){3-4} \cmidrule(lr){5-6}
& & Acc. & \#Perceived & Acc. & \#Perceived \\
\midrule
\multirow{6}{*}{\textbf{Image}} 
& \textbf{GPT-4V}              & \textbf{90.4} & \textbf{835} & \underline{86.5} & \underline{799} \\
& \textbf{Claude-3.7-Sonnet}   & \underline{88.3} & \underline{816} & \textbf{91.2} & \textbf{843} \\
& \textbf{Qwen-VL-Plus }                & 83.2 & 769 & 84.4 & 780 \\
& \textbf{Gemini-2.0-Flash }            & 86.4 & 798 & 85.4 & 789 \\
& \textbf{DeepSeek-VL }                & 56.1 & 518 & 72.9 & 674 \\
& \textbf{GLM-4V-Plus}                 & 80.1 & 740 & 84.2 & 778 \\
\midrule
\multirow{2}{*}{\textbf{Video}} 
& \textbf{LongVU }                     & 34.0 & 34 & 44.0 & 44 \\
& \textbf{Chat-UniVi}                  & 27.0 & 27 & 61.0 & 61 \\
\bottomrule
\end{tabular}
\end{table}

\subsection{Perception Capability Evaluation Results} 
Table~\ref{tab:perception} summarizes the perception-level performance of all MLMs under similarity thresholds of 0.8 for images and 0.7 for videos. These thresholds were determined through pilot studies and manual inspection to ensure that only outputs highly consistent with human-authored descriptions are counted as correct.

We evaluate two prompting strategies: standard input–output (IO) and multi-modal chain-of-thought (MCOT) prompting. For image tasks, GPT-4V achieves the highest perception accuracy under the IO setting (90.4\%), followed closely by Claude-3.7-Sonnet, Gemini-2.0-Flash, and Qwen-VL-Plus. GLM-4V-Plus performs reasonably well (80.1\%), while DeepSeek-VL lags behind at 56.1\%. In video-based perception, performance drops sharply for all models, with LongVU and Chat-UniVi achieving only 34.0\% and 27.0\% accuracy, respectively.
Switching to the MCOT prompting strategy yields slight changes. For images, Claude-3.7-Sonnet achieves the highest perception accuracy (91.2\%), surpassing GPT-4V (86.5\%). Most other models, including GLM-4V-Plus, Gemini-2.0-Flash, and Qwen-VL-Plus, also show improvements under MCOT, while DeepSeek-VL demonstrates a notable gain. For videos, both LongVU and Chat-UniVi see marked improvements (44.0\% and 61.0\%, respectively).

Notably, the end-to-end classification accuracy of LongVU, Chat-UniVi and DeepSeek-VL (see Table~\ref{tab:baseline}) exceeds their perception accuracy here. This is not a contradiction but rather a confirmation of our core hypothesis: MLMs may exploit superficial cues or dataset biases to achieve high classification scores without truly grounding their decisions in accurate perception. This finding further underscores the necessity of layered evaluation frameworks.

\begin{figure}[htp]
  \centering
  \includegraphics[width=4in]{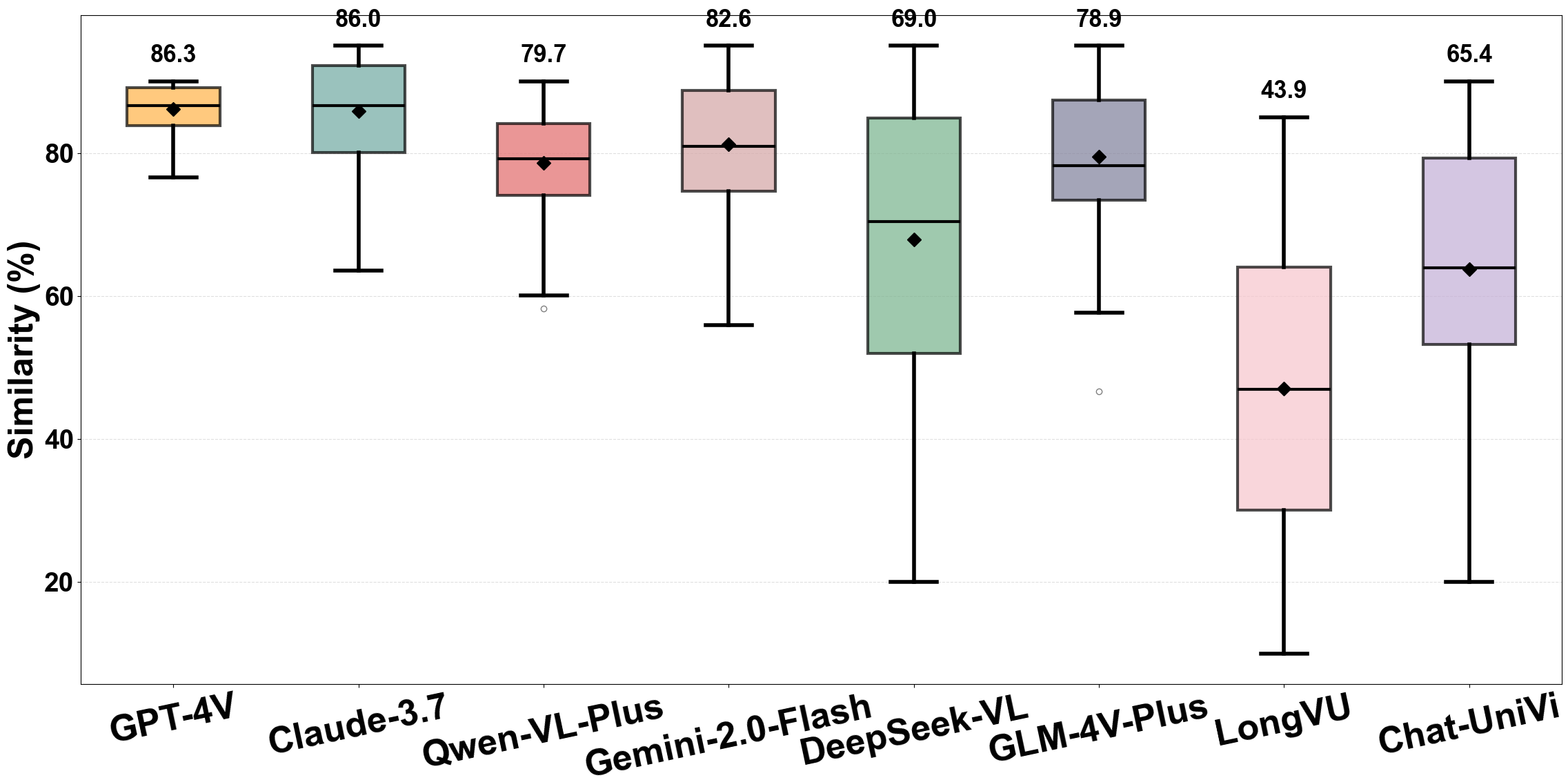}
  \caption{The scene description similarity across different MLMs under IO prompting.}
  \label{fig:similarity}
\end{figure}

In addition, we presents the distribution of scene description similarity scores in Fig.~\ref{fig:similarity}. The results shows clear performance stratification among MLMs. GPT-4V and Claude-3.7-Sonnet lead with both the highest average similarity (86.3\% and 86.0\%, respectively) and the narrowest distributions, as evidenced by their low standard deviations and high minimum values. This indicates not only superior perceptual grounding but also remarkable output stability across diverse samples. In contrast, DeepSeek-VL, LongVU, and Chat-UniVi show substantially lower averages, wider distributions, and much lower minimums, reflecting frequent failures and highly inconsistent scene perception. 
These findings emphasize that only a select few SoTA MLMs deliver both high-fidelity and reliable scene description, a prerequisite for deeper cognitive tasks. 

\begin{table}[t]
\centering
\footnotesize
\caption{The \textbf{recognition} capability evaluation results of 8 MLMs.}
\label{tab:recognition}
\begin{tabular}{llcccc}
\toprule
\multirow{2}{*}{\textbf{Modality}} & \multirow{2}{*}{\textbf{Model}}  
  & \multicolumn{2}{c}{\textbf{IO}} 
  & \multicolumn{2}{c}{\textbf{MCoT}}  \\
\cmidrule(lr){4-5} \cmidrule(lr){3-6}
&    & ~Acc.~ & ~F1~   & ~Acc.~ & ~F1~ \\
\midrule
\multirow{6}{*}{\textbf{Image}}
  & \textbf{GPT-4V}             & \underline{88.6} & \textbf{90.4}  & \underline{86.7} & 87.0 \\
  & \textbf{Claude-3.7-Sonnet}   & 87.3             & 88.8           & \textbf{88.9}    & \textbf{89.4} \\
  & \textbf{Qwen-VL-Plus}       & 80.6             & 82.9          & 80.1             & 80.6 \\
  & \textbf{Gemini-2.0-Flash}    & \textbf{89.6}    & \underline{89.9} & 84.2             & \underline{88.3} \\
  & \textbf{DeepSeek-VL}         & 72.6             & 76.1        & 66.7             & 53.8 \\
  & \textbf{GLM-4V-Plus}       & 85.0             & 88.4         & 84.5             & 87.2 \\
\midrule
\multirow{2}{*}{\textbf{Video}}
  & \textbf{LongVU}             & 28.1             & 20.7        & 42.9             & 33.3 \\
  & \textbf{Chat-UniVi}       & 48.2             & 53.2         & 59.3             & 61.8 \\
\bottomrule
\end{tabular}
\end{table}

\subsection{Cognitive Capability Evaluation Results} 
This experiment evaluates sarcasm recognition on only those samples that each model correctly perceived, as shown in Table~\ref{tab:recognition}. We are specifically measuring recognition capability under the ideal condition where perception has already succeeded.

For image-based tasks, GPT-4V and Gemini-2.0-Flash deliver the strongest performance, both in terms of accuracy and F1. GPT-4V achieves a leading F1 score of 90.4\%, with Gemini-2.0-Flash close behind at 89.9\%. Claude-3.7-Sonnet and GLM-4V-Plus also maintain high recognition scores. By contrast, DeepSeek-VL falls behind, not only does it correctly perceive fewer samples, but its recognition accuracy on those is also lower than the top models. For video-based tasks, the picture changes dramatically. Both LongVU and Chat-UniVi recognize sarcasm on very few perceptually correct samples, and their F1 scores drop sharply compared to image-based models, indicating substantial difficulty with temporal or dynamic content. Interestingly, while MCOT prompts improve perception accuracy across most MLMs (Table~\ref{tab:perception}), their impact on sarcasm recognition is quite different here. MCOT does not consistently enhance recognition performance. This pattern further supports our earlier finding: sarcasm understanding is not a purely rational, step-by-step process, and imposing linear reasoning chains may even disrupt the model's ability.

This two-stage evaluation approach reveals a crucial fact: when MLMs are able to accurately perceive images, their sarcasm recognition improves dramatically. However, for video tasks, even flawless perception does not prevent a significant drop in recognition performance. More importantly, the gap between these layered results and end-to-end classification scores highlights that strong overall predictions can sometimes mask underlying perception errors. Therefore, layered evaluation is essential for truly assessing multi-modal understanding, as it exposes hidden weaknesses that end-to-end metrics alone might overlook.

\begin{figure}[htp]
  \centering
  \includegraphics[width=3.5in]{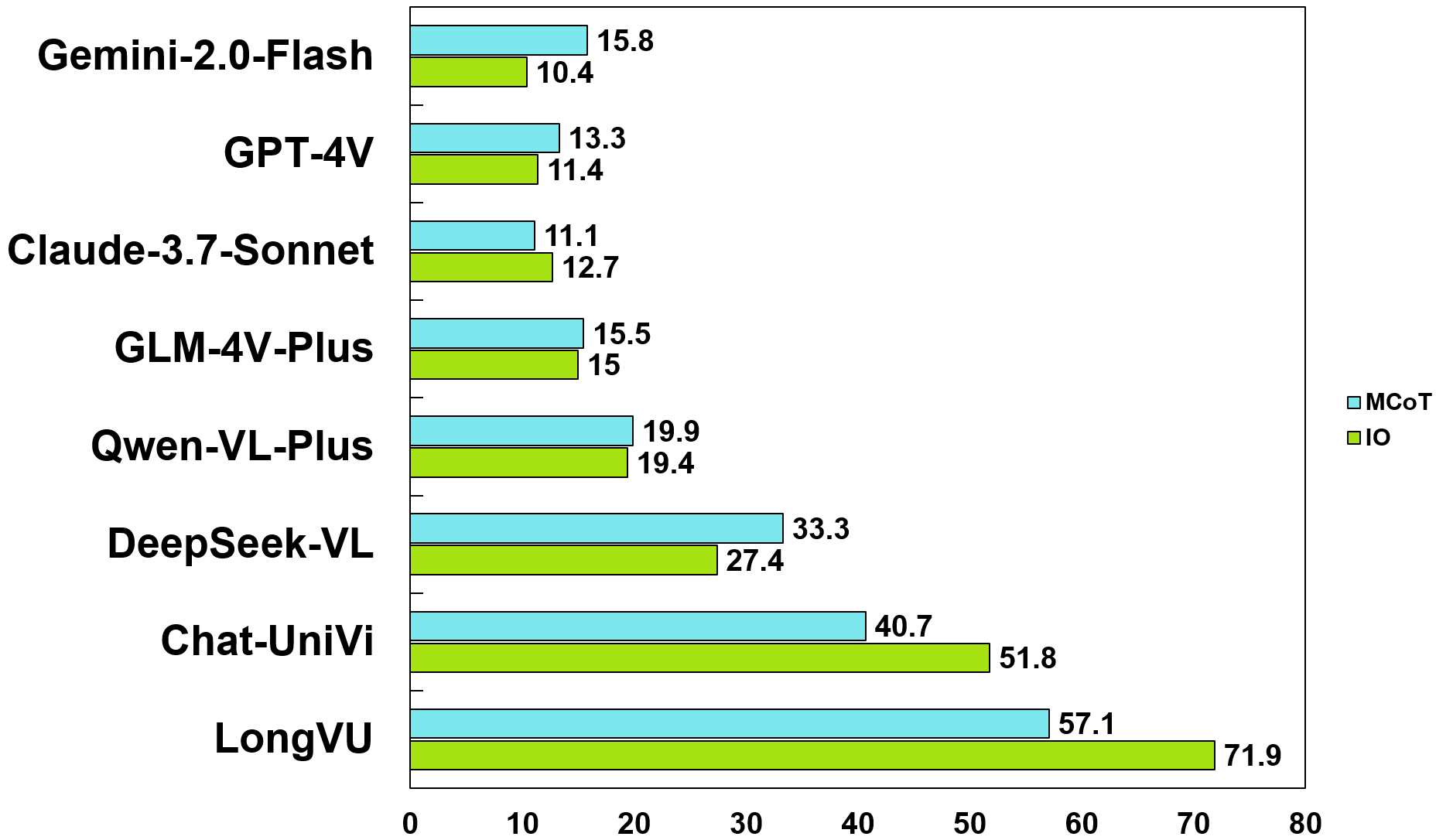}
  \caption{The distribution of gap.}
  \label{fig:gap2}
\end{figure}

\subsection{Perception–Cognition Gap Analysis} 
We report the perception–cognition gap, defined as the failure rate on sarcasm recognition among samples that have already passed the perception stage, as shown in Fig.~\ref{fig:gap2}. This metric provides a straightforward illustration of the central dilemma highlighted by our Visual Room argument: accurate perception does not guarantee genuine understanding.

For image tasks, leading MLMs such as GPT-4V, Gemini-2.0-Flash, and Claude-3.7-Sonnet exhibit relatively small gaps, with error rates ranging from 10.4\% to 12.7\%. Qwen-VL-Plus and GLM-4V-Plus show somewhat higher gaps, while DeepSeek-VL's gap reaches 27.4\%. On average, the perception–cognition gap for images is 17.1\% across two prompting methods, indicating that even SoTA MLMs still fail to achieve correct semantic understanding in a notable fraction of cases, despite having already perceived the scene accurately. For video tasks, the perception–cognition gap increases dramatically. LongVU and Chat-UniVi exhibit gaps as high as 51.8\% and 71.9\% via IO prompting, respectively, with the average gap for videos reaching 55.4\%. Compared to image tasks, this sharp disparity shows that even when MLMs can successfully recognize surface content in video, their ability to grasp deeper semantic or contextual sarcasm remains severely limited. 

In summary, these findings provide strong empirical support for our central claim: even with perfect perception, current MLMs remain far from genuine semantic understanding. The issue is especially pronounced for video tasks. This demonstrates that existing MLMs are still largely confined to surface-level, ``Visual Room''-style symbolic processing and have yet to achieve deep integration of perception and cognition. 



\subsection{Error Analysis} 
\begin{figure}[t]
  \centering
  \includegraphics[width=3.3in]{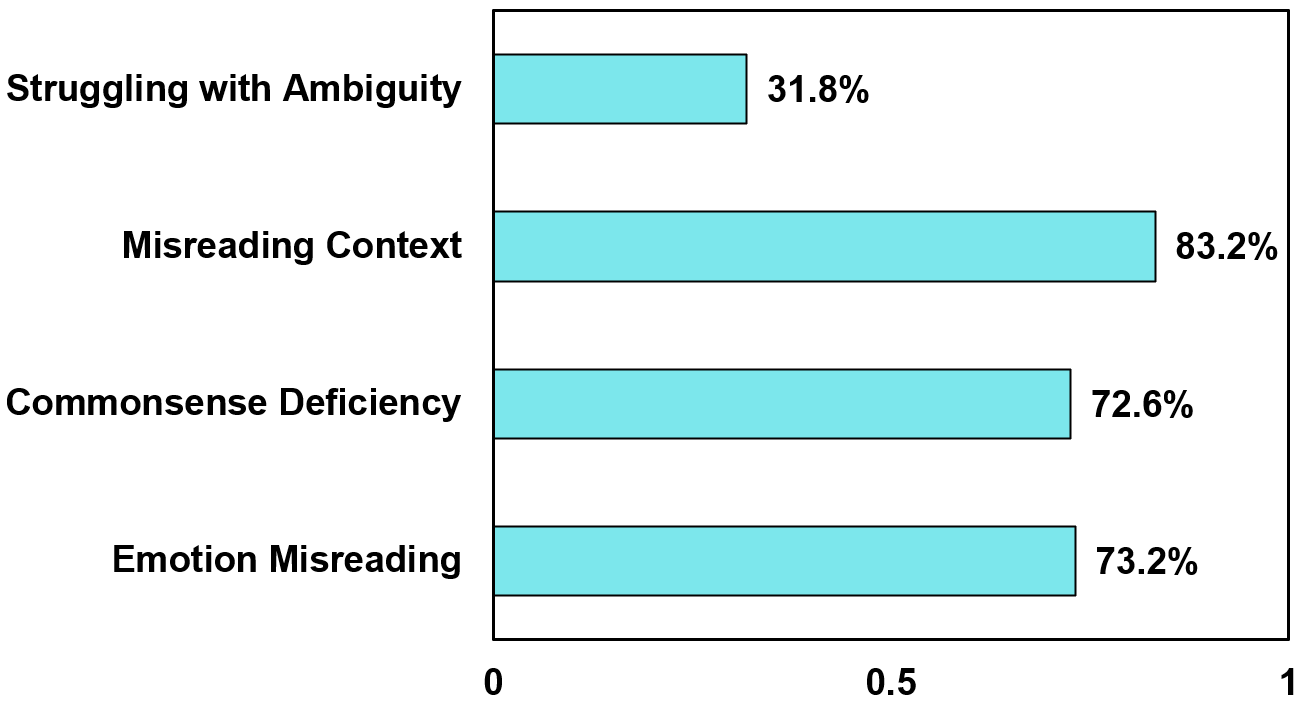}
  \caption{Error analysis.}
  \label{fig:erroranalysis}
\end{figure}

We conduct a detailed analysis of GPT-4V's errors in sarcasm recognition, as shown in Fig.~\ref{fig:erroranalysis}. We find that the most frequent type of mistake is misinterpreting the context (83.2\%), followed by misunderstanding emotions (73.2\%) and lacking commonsense knowledge (72.6\%). These results suggest that GPT-4V, even when correctly recognizing visual details, often fails to grasp subtle situational cues, emotions, or ironic intentions, instead providing superficial or literal answers. Additionally, the model struggles with ambiguity in 31.8\% of the cases, producing unclear or inconsistent responses when faced with uncertain input.

Overall, these error patterns strongly align with our Visual Room argument: even SoTA MLMs primarily perform symbolic manipulation, still lacking deep and context-sensitive understanding.
\begin{figure}[t]
  \centering
  \includegraphics[width=4.8in]{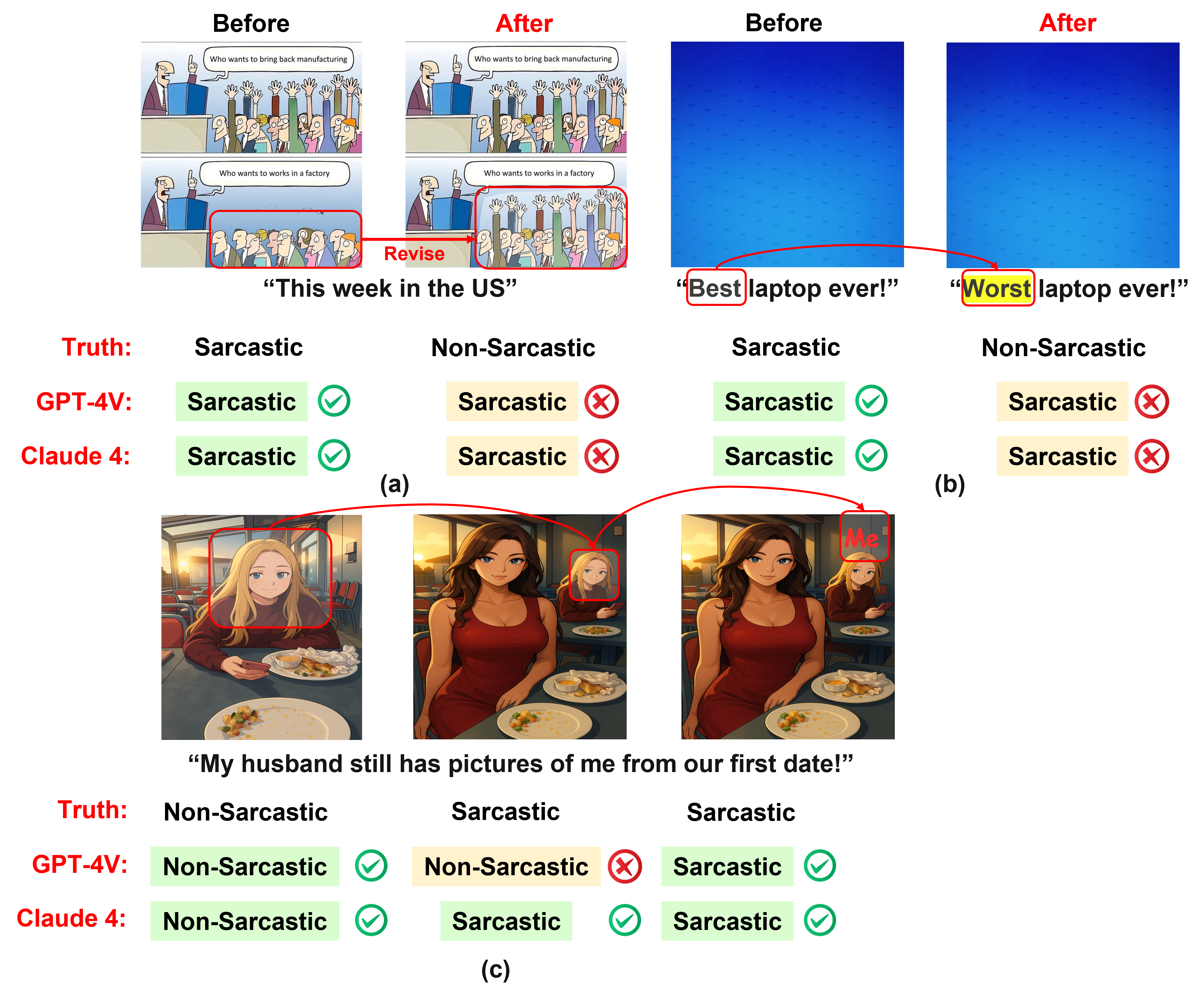}
  \caption{Counterfactual reasoning results.}
  \label{fig:Counterfactual222}
\end{figure}

\subsection{Counterfactual Reasoning Analysis}
We conduct a counterfactual analysis, as illustrated in Fig.~\ref{fig:Counterfactual222}. Specifically, we slightly revise critical visual or textual details of representative examples to intentionally change the intended meaning and thus re-evaluate the MLMs' predictions.
In Fig.~\ref{fig:Counterfactual222}(a), we modify the lower panel so that all audience members raise their hands, showing they are willing to work in factories. This shifts the image's original sarcastic meaning to a non-sarcastic one. However, GPT-4V and Claude 4 fail to recognize the change and still interpret the image as sarcastic. 
In Fig~\ref{fig:Counterfactual222} (b), we change the caption from ``Best laptop ever!'' to ``Worst laptop ever!'' for an image of a blue-screened laptop. This turns the sarcasm into direct criticism, which is not sarcasm. Yet both GPT-4V and Claude 4 still classify the image as sarcastic. 

Fig.~\ref{fig:Counterfactual222}(c) starts with a non-sarcastic image of a woman dining, captioned ``My husband still has pictures of me from our first date!''. Both models correctly identify the original intent. We then modify the image by inserting a prominently placed attractive woman and placing the protagonist in a small corner, making the message sarcastic. GPT-4V misidentifies the new woman as the speaker and still predicts non-sarcastic. Claude 4 identifies the sarcasm but gives an error explanation, relying on superficial image style rather than real understanding. Only after adding the label ``Me'' above the true protagonist do both models correctly detect and explain the sarcasm, pointing to the strong contrast and self-deprecating tone.

Their failure to track subtle context shifts in counterfactual scenarios reveals a clear perception–cognition gap.


\subsection{Ablation Study}
We perform an ablation study to examine how different input components affect model performance. Table~\ref{tab:ablated} reports results for three configurations: without vision, without text, and without scene description.

Compared to the full IO setting (Table~\ref{tab:baseline}), all models show clear performance drops when either vision or text is removed. For instance, GPT-4V's F1 drops from 86.7 to 77.9 without vision, and to 61.1 without text, confirming the strong complementary role of both modalities. Claude-3.7-Sonnet and Gemini-2.0-Flash show even larger declines in the w/o Text condition, highlighting their heavy reliance on language cues.

Removing the scene description results in a moderate performance decrease. GPT-4V's F1 drops from 86.7 to 80.2, and similar trends are observed across other models. This suggests that explicit scene grounding improves cognitive interpretation but is not a substitute for vision or text. Video-based models (LongVU, Chat-UniVi) perform poorly in all ablation settings, especially when vision is removed. This reflects weaker visual understanding and highlights their dependency on shallow textual associations.

In summary, the results demonstrate that all three components, i.e., vision, text, and scene grounding play distinct and complementary roles in effective sarcasm recognition.

\begin{table}[ht]
\caption{The ablated experiment.}
\label{tab:ablated}
\centering
\begin{tabular}{lcccccc}
\hline
\multirow{2}{*}{\textbf{Model}}   & \multicolumn{2}{c}{\textbf{w/o vision}} & \multicolumn{2}{c}{\textbf{w/o text}} & \multicolumn{2}{c}{\textbf{w/o scene}} \\
               & Acc. & F1 & Acc. & F1   & Acc.    & F1     \\ \toprule
\textbf{GPT-4V}         & 76.6     & 77.9   & 66.5     & 61.1     & 82.1         & 80.2        \\
\textbf{Claude-3.7-Sonnet} & 80.1  & 77.8   & 59.8     & 41.6     & 79.8         & 76.8        \\
\textbf{Qwen-VL-Plus }  & 74.0     & 72.5   & 66.0     & 62.3     & 80.4         & 78.3        \\
\textbf{Gemini-2.0-Flash} & 79.3   & 77.3   & 59.2     & 42.8     & 81.1         & 79.4        \\
\textbf{DeepSeek-VL}    & 68.6     & 62.7   & 68.6     & 63.7     & 76.7         & 75.6        \\
\textbf{GLM-4V-Plus}    & 79.7     & 78.5   & 71.3     & 71.3     & 82.0         & 81.3        \\ \toprule

\textbf{LongVU} & 43.0     &32.9    &50.0      &54.5      &59.0          &70.1         \\
\textbf{Chat-UniVi}         &62.2      &65.9    &48.5      &61.7      &56.1          &69.1        \\ \toprule
\end{tabular}

\end{table}

\begin{table}[t]
\caption{The experimental results on multi-modal humor understanding.}
\label{tab:Transferability}
\centering
\begin{tabular}{lcccc}
\hline
\textbf{Model} & \textbf{Perception}                               & \textbf{Recognition} & \textbf{Gap} \\ \toprule
\textbf{GPT-4V}           & \textbf{85.0}                           & \textbf{88.2}                          & \textbf{11.8}                          \\
\textbf{Claude-3.7-Sonnet} & 74.0                           & 87.8                          & 12.2                          \\ \toprule
\end{tabular}
\end{table}

\subsection{Paradigm Transferability Analysis} 
To demonstrate the generalizability of our framework, we apply it to a related cognitive task: humor understanding. Like sarcasm, humor involves non-literal expression, pragmatic reasoning, and context-sensitive interpretation. We collect 100 samples (50 humorous, 50 non-humorous) from Reddit's ``r/humor'' subreddit, where each instance is self-labeled by the original author.

As shown in Table~\ref{tab:Transferability}, GPT-4V and Claude-3.7-Sonnet both perform well in this task. GPT-4V achieves a perception accuracy of 85.0\% and recognition accuracy of 88.2\%, with a perception–cognition gap of 11.8\%. Claude-3.7-Sonnet shows a similar gap (12.2\%) despite lower perception accuracy. These results suggest that our two-stage evaluation framework generalizes well beyond sarcasm and can be effectively applied to other nuanced multi-modal reasoning tasks.


\section{Conclusion}
\label{sec:conclusion}
In this paper, we propose the Visual Room argument to highlight the gap between perception and genuine understanding in MLMs. Through a two-stage evaluation framework and a new sarcasm dataset (MMSar), we show that even SoTA MLMs often fail to comprehend intent, especially in non-literal tasks like sarcasm and humor. Our experiments reveal consistent perception–cognition gaps, misleading end-to-end scores, and reasoning failures rooted in shallow symbol manipulation. These findings confirm that current MLMs, though strong in surface perception, remain far from achieving true semantic understanding.

\section{Limitations}
\label{sec:limitation_ethic}
\textbf{Limitations.} This study has several limitations. First, it focuses only on sarcasm and humor, without exploring other types of cognitive reasoning. Second, the dataset size is relatively small, especially for videos, which may limit generalizability. Third, we evaluate models in zero-shot settings only, without exploring methods to reduce the perception–cognition gap through fine-tuning. Fourth, while the Visual Room framework is well-supported, we do not examine how it could be improved or adapted. Finally, the counterfactual analysis is based on a few illustrative examples, lacking broader-scale quantification.

\textbf{Potential risks.} Counterfactual image editing, while used here to evaluate model capabilities, could be misused for disinformation or manipulation such as altering sarcastic contexts to distort intent, spread misleading narratives, or launch reputational attacks.

\section*{Acknowledgements}
Thank all the anonymous reviewers and chairs for their constructive suggestions.

%
%
%
 \bibliographystyle{splncs04}
 \bibliography{mybibliography}

\end{document}